\documentclass[11pt,a4paper]{article}
\usepackage[hyperref]{acl2019}
\usepackage{times}
\usepackage{latexsym}
\usepackage{helvet}  
\usepackage{courier}  
\usepackage{url}  
\usepackage{graphicx}  
\usepackage{booktabs} 
\usepackage{tabularx}
\usepackage{arydshln}
\usepackage{wrapfig}
\usepackage{amsmath}
\usepackage{amsfonts}
\usepackage{graphicx}
\usepackage{subcaption}
\usepackage{algorithm}
\usepackage{algorithmic}
\usepackage{latexsym}
\usepackage{multirow}
\usepackage{multirow}
\usepackage{tablefootnote}

\usepackage{url}

\aclfinalcopy 

\title{Multi-turn Dialogue Response Generation in an \\Adversarial Learning Framework}

\author{Oluwatobi Olabiyi \\
  Capital One Conversation Research \\
  Vienna VA \\
  \textit{oluwatobi.olabiyi@capitalone.com} \\\And
  Alan Salimov \\
  Capital One Conversation Research \\
  San Francisco CA \\
  \textit{ alan.salimov@capitalone.com} \\\AND
  Anish Khazane \\
  Capital One Conversation Research \\
  San Francisco CA \\
  \textit{ anish.khazane@capitalone.com} \\\And
  Erik T. Mueller \\
  Capital One Conversation Research \\
  Vienna VA \\
  \textit{ erik.mueller@capitalone.com} \\}

\date{}

\begin{document}
\maketitle
\begin{abstract}
  We propose an adversarial learning approach for generating multi-turn dialogue responses.
  Our proposed framework, \textit{hredGAN}, is based on conditional generative adversarial
  networks (GANs). The GAN's generator is a modified hierarchical recurrent encoder-decoder network (HRED) 
  and the discriminator is a word-level bidirectional RNN that shares context and word embeddings with the generator. 
  During inference, noise samples conditioned on the dialogue history are used to perturb
  the generator's latent space to generate several possible responses. 
  The final response is the one ranked best by the discriminator. The hredGAN shows improved performance 
  over existing methods: (1) it generalizes better than 
  networks trained using only the log-likelihood criterion, and (2) it generates longer, 
  more informative and more diverse responses with high utterance and topic relevance even 
  with limited training data. This improvement is demonstrated on the Movie triples and
  Ubuntu dialogue datasets using both automatic and human evaluations.
\end{abstract}



\section{Introduction}
\label{introduction}
Recent advances in deep neural network architectures have enabled
tremendous success on a number of difficult machine learning problems.
While these results are impressive, producing
a deployable neural network--based model that can
engage in open domain conversation still remains elusive.
A dialogue system needs to be able to generate meaningful and
diverse responses that are simultaneously coherent with the input utterance
and the overall dialogue topic.
Unfortunately, earlier conversation models trained with naturalistic
dialogue data suffered greatly from limited contextual information 
\citep{Sutskever2014,Vinyals2015} and lack of diversity \citep{Li2016}. 
These problems often lead to generic and 
safe responses to a variety of input utterances. 

\citet{Serban2016} and \citet{Xing} proposed the Hierarchical Recurrent Encoder-Decoder 
(HRED) network to capture long temporal dependencies in multi-turn conversations to 
address the limited contextual information but the diversity problem 
remained. In contrast, some HRED variants such as
variational \citep{Serban2017} and multi-resolution \citep{Serban2017a}
HREDs attempt to alleviate the diversity problem by injecting noise
at the utterance level and by extracting additional context to condition
the generator on. While these approaches achieve a certain
measure of success over the basic HRED, the generated responses are still mostly generic 
since they do not control the generator's output. This is because the output conditional distribution is not calibrated. 
\citet{Li2016}, on the other hand, 
consider a diversity promoting training objective but their model is for single turn conversations and 
cannot be trained end-to-end. 

The generative adversarial network (GAN) \citep{Goodfellow2014} seems to
be an appropriate solution to the diversity problem. GAN
matches data from two different distributions by introducing an
adversarial game between a \textit{generator} and a \textit{discriminator}. 
We explore \textit{hredGAN}: conditional
GANs for multi-turn dialogue models with an HRED generator and discriminator.
hredGAN combines ideas from both generative and retrieval-based multi-turn dialogue systems to 
improve their individual performances. This is achieved by sharing the context and 
word embeddings between the generator and the discriminator allowing for 
joint end-to-end training using back-propagation.  
To the best of our knowledge, no existing work has applied conditional GANs
to multi-turn dialogue models and especially not with
HRED generators and discriminators. We demonstrate the effectiveness of hredGAN
over the VHRED for dialogue modeling with evaluations on the
Movie triples and Ubuntu technical support datasets.

\section{Related Work}

Our work is related to end-to-end neural network--based open domain
dialogue models. Most neural dialogue models use transduction
frameworks adapted from neural machine translation \citep{Sutskever2014,Bahdanau2015}. 
These \texttt{Seq2Seq} networks are trained end-to-end
with MLE criteria using large corpora of human-to-human conversation
data. 
Others use GAN's discriminator as a reward function in a reinforcement
learning framework \citep{Yu2017} and in conjunction with MLE
\citep{Li2017,Che2017}. \citet{Zhang2017} explored
the idea of GAN with a feature matching criterion. \citet{Xu2017} and \citet{Zhang2018b} employed 
GAN with an approximate embedding layer as well as with adversarial information maximization 
respectively to improve \texttt{Seq2Seq}'s diversity performance.

Still, \texttt{Seq2Seq} models are limited in their ability to
capture long temporal dependencies in multi-turn conversation. Although \citet{Li2016c} 
attempted to optimize a pair of \texttt{Seq2Seq} models for multi-turn dialogue, the multi-turn 
objective is only applied at inference and not used for actual model training. 
Hence the introduction of HRED models \citep{Serban2016,Serban2017a,Serban2017,Xing} 
for modeling dialogue response in multi-turn
conversations. However, these HRED models suffer from lack of diversity
since they are only trained with MLE criteria. On the other hand, adversarial system has been 
used for evaluating open domain dialogue models \citep{Bruni2017,Kannan2017}.
Our work, hredGAN, is closest to the combination of HRED generation models
\citep{Serban2016} and adversarial evaluation \citep{Kannan2017}.

\section{Model}

\subsection{Adversarial Learning of Dialogue Response}
\label{adv_learn}

Consider a dialogue consisting of a sequence of $N$ utterances,
$\boldsymbol{x}=\big(x_1,x_2,\cdots,x_N\big)$, where each utterance
$x_i=\big(x_i^1,x_i^2,\cdots,x_i^{M_i }\big)$ contains a
variable-length sequence of $M_i$ word tokens such that ${x_i}^j \in V $ for
vocabulary $V$. At any time step $i$, the dialogue history is given
by $\boldsymbol{x_i}=\big(x_1,x_2,\cdots,x_i\big)$. The dialogue
response generation task can be defined as follows: Given a dialogue history
$\boldsymbol{x_i}$, generate a response
$y_i=\big(y_i^1,y_i^2,\cdots,y_i^{T_i}\big)$, where $T_i$ is the
number of generated tokens. We also want the distribution of the generated
response $P(y_i)$ to be indistinguishable from that of the ground truth
$P(x_{i+1})$ and $T_i=M_{i+1}$. 
Conditional GAN learns a mapping from an observed
dialogue history, $\boldsymbol{x_i}$, and a sequence of random noise
vectors, $z_i$ to a sequence of output tokens, $y_i$,
$G:\{\boldsymbol{x_i},z_i\} \rightarrow y_i$. The generator $G$ is
trained to produce output sequences that cannot be distinguished
from the ground truth sequence by an adversarially trained discriminator
$D$ that is trained to do well at detecting the generator's fakes.
The distribution of the generator output sequence can be factored by the product rule:
\begin{align}
P(y_i|\boldsymbol{x_i}) = P(y_i^1)\prod_{j=2}^{T_i}P\big(y_i^j | y_i^1,\cdots,y_i^{j-1}, \boldsymbol{x_i}\big)
\label{eq:pf}
\end{align}
\begin{align}
P\big(y_i^j | y_i^1,\cdots,y_i^{j-1}, \boldsymbol{x_i}\big) = P_{\theta_G}\big(y_i^{1:j-1}, \boldsymbol{x_i}\big)
\label{eq:pf}
\end{align}
where $y_i^{i:j-1} = (y_i^1,\cdots,y_i^{j-1})$ and $\theta_G$ are the
parameters of the generator model.
$P_{\theta_G}\big(y_i^{i:j-1}, \boldsymbol{x_i}\big)$ is an autoregressive generative
model where the probability of the current token depends on the past generated sequence. 
Training the generator $G$ with the log-likelihood criterion is unstable in practice, and therefore the past generated sequence
is substituted with the ground truth, a method known as \textit{teacher forcing} \citep{Williams1989}, i.e.,
\begin{equation}
P\big(y_i^j | y_i^1,\cdots,y_i^{j-1}, \boldsymbol{x_i}\big) \approx P_{\theta_G}\big(x_{i+1}^{1:j-1}, \boldsymbol{x_i}\big)
\label{eq:tf}
\end{equation}
Using \eqref{eq:tf} in relation to GAN, we define our fake sample as the teacher forcing output with some input noise $z_i$
\begin{equation}
y_i^j \sim P_{\theta_G}\big(x_{i+1}^{1:j-1}, \boldsymbol{x_i}, z_i\big)
\label{eq:sample}
\end{equation}
and the corresponding real sample as ground truth $x_{i+1}^j$.

With the GAN objective, we can match the noise distribution, $P(z_i)$, to the distribution of the ground truth response, $P(x_{i+1}|\boldsymbol{x_i})$. 
Varying the noise input then allows us to generate diverse responses to the same dialogue history. Furthermore, the discriminator, 
since it is calibrated, is used during inference to rank the generated responses, providing a means of controlling the generator output.

\subsubsection{Objectives}

The objective of a conditional GAN can be expressed as
\begin{multline} \label{eq:cgan}
\mathcal{L}_{cGAN}(G,D) = \mathbb{E}_{\boldsymbol{x_i},x_{i+1}}[\log~D(x_{i+1}, \boldsymbol{x_i})] + \\
\mathbb{E}_{\boldsymbol{x_i},z_i}[1-\log{}D(G(\boldsymbol{x_i},z_i), \boldsymbol{x_i})]
\end{multline}
where $G$ tries to minimize this objective against an adversarial $D$ that tries to maximize it:
\begin{equation}
G^*, D^* = arg\mathop{min}\limits_G\mathop{max}\limits_D\mathcal{L}_{cGAN}(G,D).
\end{equation}
Previous approaches have shown that it is beneficial to mix the GAN objective with 
a more traditional loss such as cross-entropy loss \citep{Lamb2016,Li2017}.
The discriminator's job remains unchanged, but the generator is tasked not 
only to fool the discriminator but also to be near the ground truth $x_{i+1}$ in the cross-entropy sense:
\begin{equation} \label{eq:mle}
\mathcal{L}_{MLE}(G) = \mathbb{E}_{\boldsymbol{x_i},x_{i+1},z_i}[-log~P_{\theta_G}\big(x_{i+1}, \boldsymbol{x_i}, z_i\big)].
\end{equation}
Our final objective is, 
\begin{multline} \label{eq:mlegan}
G^*, D^* = arg\mathop{min}\limits_G\mathop{max}\limits_D\big(\lambda_{G}\mathcal{L}_{cGAN}(G,D) + \\
\lambda_{M}\mathcal{L}_{MLE}(G)\big).
\end{multline}

\begin{figure*}[t]
\begin{center}
\begin{subfigure}{.47\textwidth}
\centerline{\includegraphics[width=\textwidth]{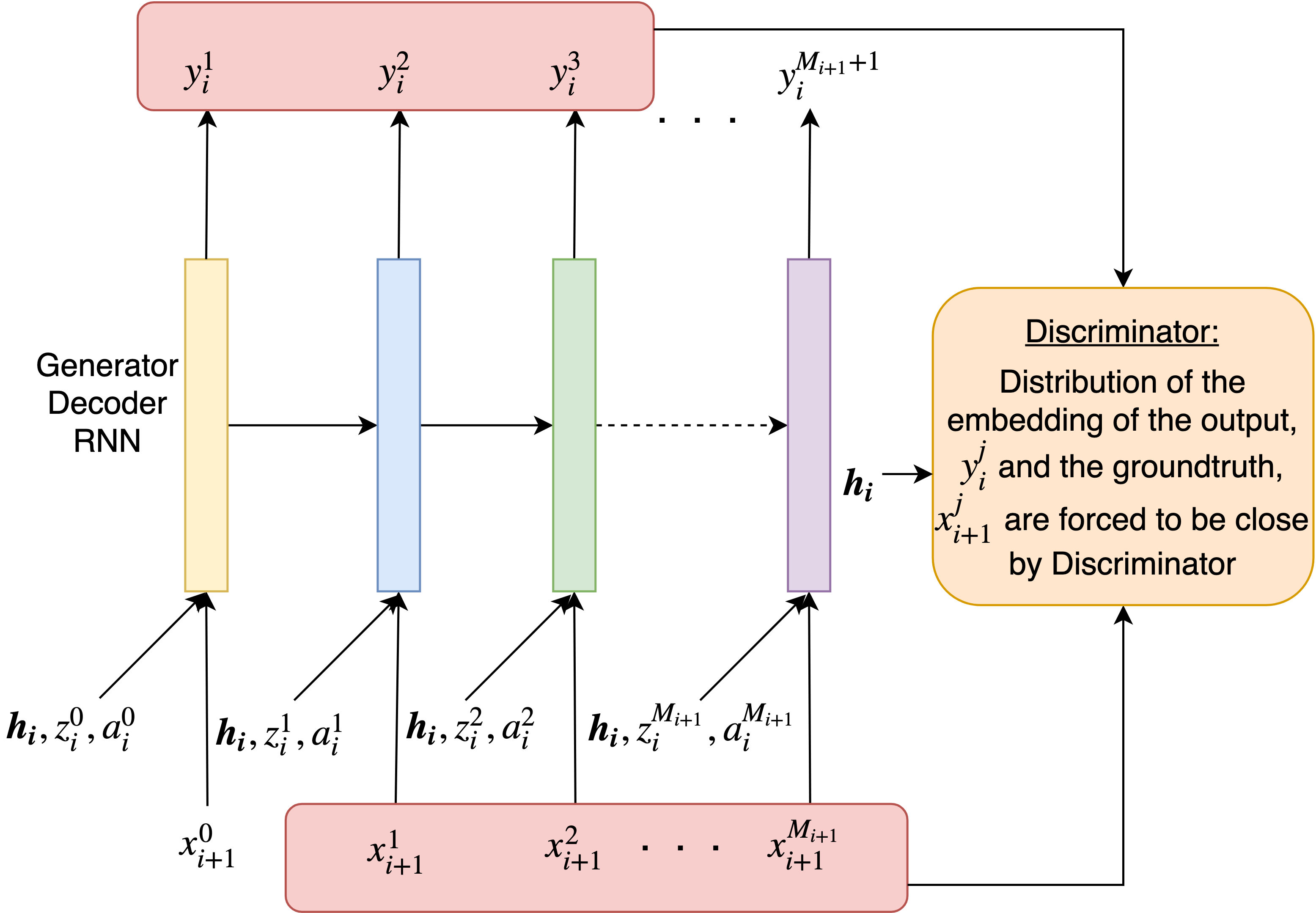}}

\end{subfigure}%
\hfill
\begin{subfigure}{.40\textwidth}
\centerline{\includegraphics[width=\textwidth]{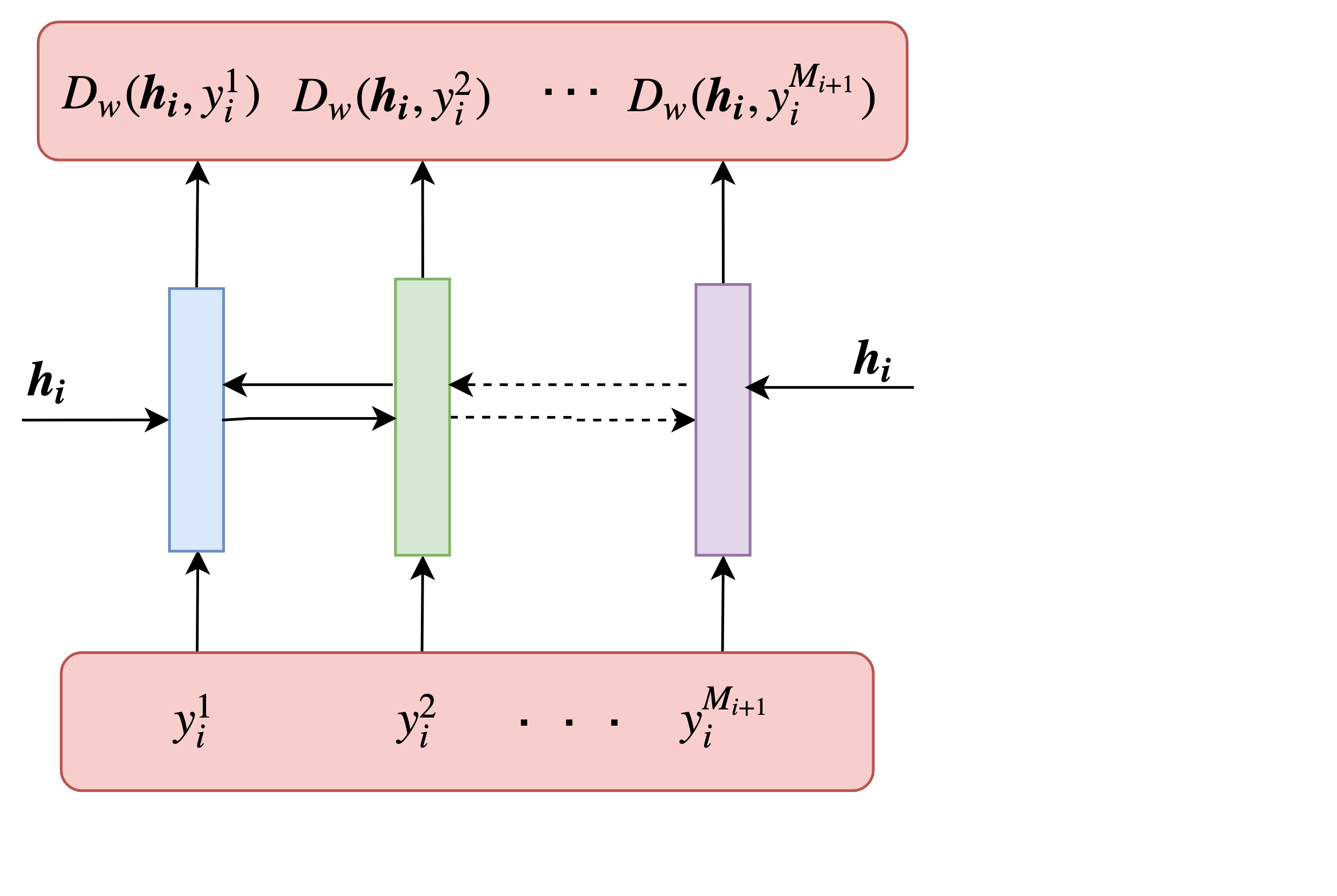}}
\end{subfigure}
\caption{\textbf{Left: The hredGAN architecture -} The generator makes predictions conditioned on 
the dialogue history, $\boldsymbol{h_i}$, attention, $a_i^j$, noise sample, $z_i^j$, and ground truth, $x_{i+1}^{j-1}$.
 \textbf{Right: RNN-based discriminator} that discriminates bidirectionally at the word level.}
 \label{hred_gan}
\end{center}
\vskip -0.2in
\end{figure*}

It is worth mentioning that, without $z_i$, the net could still learn a mapping 
from $\boldsymbol{x_i}$ to $y_i$, 
but it would produce deterministic outputs and fail to match any distribution other than a delta function \citep{Isola2017}.
This is one key area where our work is different from Lamb et~al.'s and Li et~al.'s.
The schematic of the proposed hredGAN is depicted at the right hand side of Figure~\ref{hred_gan}.

\subsubsection{Generator}
We adopted an HRED dialogue generator similar to \citet{Serban2016,Serban2017a,Serban2017} and \citet{Xing}.
The HRED contains three recurrent structures, i.e. 
the encoder $(eRNN)$, context $(cRNN)$, and decoder $(dRNN)$ RNN.
The conditional probability modeled by the HRED per output word token is given by
\begin{equation} \label{eq:hred}
P_{\theta_G}\big(y_i^j|x_{i+1}^{1:j-1},\boldsymbol{x_i}\big) = dRNN\big(E(x_{i+1}^{j-1}), h_i^{j-1}, \boldsymbol{h_i}\big)
\end{equation}
where $E(.)$ is the embedding lookup, $\boldsymbol{h_i} = cRNN(eRNN(E(x_i),\boldsymbol{h_{i-1}})$, $eRNN(.)$ maps a sequence of 
input symbols into fixed-length vector, and $h$ and $\boldsymbol{h}$ are the hidden states of the decoder and context RNN, respectively.

\begin{figure*}[ht]
\begin{center}
\centerline{\includegraphics[width=0.85\textwidth]{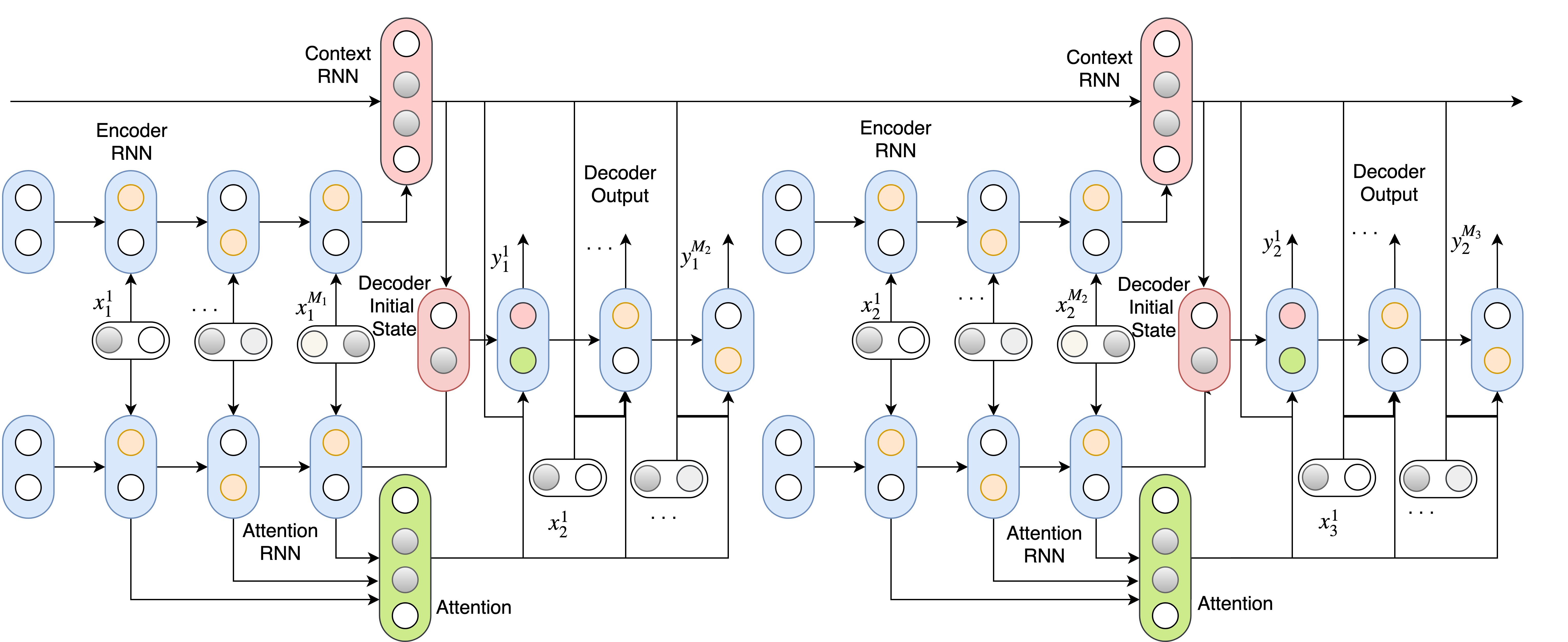}}
\caption{\textbf{The HRED generator with local attention -}
The attention RNN ensures local relevance while the context RNN ensures global relevance. 
Their states are combined to initialize the decoder RNN and the discriminator BiRNN.}
\label{hred_att}
\end{center}
\vskip -0.2in
\end{figure*}

In the multi-resolution HRED, \citep{Serban2017a}, high-level tokens are extracted and processed by another RNN to improve 
performance. We circumvent the need for this extra processing by allowing the decoder to attend to different parts of the input 
utterance during response generation \citep{Bahdanau2015,Luong2015}.
We introduce a local attention into \eqref{eq:hred} and 
encode the attention memory differently from the context through an attention encoder RNN $(aRNN)$, yielding:
\begin{multline} \label{eq:ahred}
P_{\theta_G}\big(y_i^j|x_{i+1}^{1:j-1},\boldsymbol{x_i}\big) = \\ dRNN\big(E(x_{i+1}^{j-1}), h_i^{j-1}, a_i^j, \boldsymbol{h_i}\big)
\end{multline}
where $a_i^j = \sum_{m=1}^{M_i}\frac{exp(\alpha_m)}{\sum_{m=1}^{M_i}exp(\alpha_m)}h_i^{'m},$
$h_i^{'m} = aRNN(E(x_i^m), h_i^{'m-1})$, $h^{'}$ is the hidden state of the attention RNN, and $\alpha_k$ is 
either a logit projection of $(h_i^{j-1}, h_i^{'m})$ in the case of \citet{Bahdanau2015} or $(h_i^{j-1})^T\cdot h_i^{'m}$ in the case of \citet{Luong2015}.
The modified HRED architecture is shown in Figure~\ref{hred_att}.

\textbf{Noise Injection:} We inject Gaussian noise at the input of the decoder RNN.
Noise samples could be injected at the utterance or word level.
With noise injection, the conditional probability of the decoder output becomes
\begin{multline} \label{eq:nahred}
P_{\theta_G}\big(y_i^j|x_{i+1}^{1:j-1},z_i^j,\boldsymbol{x_i}\big) =  \\ 
dRNN\big(E(x_{i+1}^{j-1}), h_i^{j-1}, a_i^j, z_i^j, \boldsymbol{h_i}\big)
\end{multline}
where $z_i^j \sim \mathcal{N}_i(0,\boldsymbol{I})$, for utterance-level noise 
and $ z_i^j \sim \mathcal{N}_i^j(0,\boldsymbol{I})$, for word-level noise.

\subsubsection{Discriminator}

The discriminator shares context and word embeddings with the generator and can discriminate at the word level \citep{Lamb2016}.
The word-level discrimination is achieved through a bidirectional RNN and is able to capture both 
syntactic and conceptual differences between the generator output and the ground truth.
The aggregate classification of an input
sequence, $\chi$ can be factored over word-level discrimination and expressed as
\begin{align} \label{eq:dis}
D(\boldsymbol{x_i},\chi) = D(\boldsymbol{h_i},\chi)
= \bigg[\prod_{j=1}^{J} D_{RNN}(\boldsymbol{h_i},E(\chi^j))\bigg]^{\frac{1}{J}}
\end{align}
where $D_{RNN}(.)$ is the word discriminator RNN, $\boldsymbol{h_i}$ is an encoded vector of the dialogue history 
$\boldsymbol{x_i}$ obtained from the generator's $cRNN(.)$ output, and $\chi^j$ is the \textit{jth} 
word or token of the input sequence $\chi$.
$\chi = y_i$ and $J=T_i$ for the case of generator's decoder output, $\chi = x_{i+1}$ and $J=M_{i+1}$ 
for the case of ground truth.
The discriminator architecture is depicted on the left hand side of Figure~\ref{hred_gan}.

\subsection{Adversarial Generation of Multi-turn Dialogue Response}
\label{adv_gen} \abovedisplayskip=0pt \belowdisplayskip=0pt
In this section, we describe the generation process during inference. 
The generation objective can be mathematically described as 
\begin{equation} \label{eq:adv_gen}
y_i^* = arg\mathop{max}\limits_l\big\{P(y_{i,l}|\boldsymbol{x_i}) + D^*(\boldsymbol{x_i}, y_{i,l})]\big\}_{l=1}^{L}
\end{equation}
where $y_{i,l} = G^*(\boldsymbol{x_i},z_{i,l})$, $z_{i,l}$ is the \textit{lth} 
noise samples at dialogue step $i$, and $L$ is the number of response samples. 
Equation \ref{eq:adv_gen} shows that our inference objective is the same 
as the training objective (\ref{eq:mlegan}), combining both the MLE and 
adversarial criteria. This is in contrast to existing work where the discriminator 
is usually discarded during inference.

The inference described by \eqref{eq:adv_gen} is intractable due 
to the enormous search space of $y_{i,l}$. Therefore, we turn to an approximate 
solution where we use greedy decoding (MLE) on the first part of the objective
function to generate $L$ lists of responses based on noise samples $\{z_{i,l}\}_{l=1}^{L}$. In order
to facilitate the exploration of the generator's latent space, we sample a modified noise distribution,
$z_{i,l}^j \sim \mathcal{N}_{i,l}(0,\alpha\boldsymbol{I})$, or  $z_{i,l}^j \sim \mathcal{N}_{i,l}^j(0,\alpha\boldsymbol{I})$ where
$\alpha > 1.0$, is the exploration factor that increases the noise variance.
We then rank the $L$ lists using the discriminator score, \big\{$D^*(\boldsymbol{x_i}, y_{i,l})]\big\}_{l=1}^{L}$.
The response with the highest discriminator ranking is the optimum response for the dialogue
context.

\begin{algorithm}[t]
  \scriptsize
\begin{algorithmic}
   \label{alg:hredgan}
   \REQUIRE A generator $G$ with parameters $\theta_G$. 
   \REQUIRE A discriminator $D$ with parameters $\theta_D$.
   \FOR {number of training iterations}
   \STATE Initialize $cRNN$ to zero\_state, $\boldsymbol{h_0}$
   \STATE Sample a mini-batch of conversations, $\boldsymbol{x} = \{x_i\}_{i=1}^N$, 
   $\boldsymbol{x_i} = (x_1, x_2, \cdots, x_i)$ with $N$ utterances. 
   Each utterance mini batch $i$ contains $M_i$ word tokens.
   \FOR{$i=1$ {\bfseries to} $N-1$}
   \STATE Update the context state.
   \STATE $\boldsymbol{h_i} = cRNN(eRNN(E(x_i)),\boldsymbol{h_{i-1}})$
   \STATE Compute the generator output using \eqref{eq:nahred}.
   \STATE $P_{\theta_G}\big(y_i|,z_i,\boldsymbol{x_i}\big) = \big\{P_{\theta_G}\big(y_i^j|x_{i+1}^{1:j-1},z_i^j,\boldsymbol{x_i}\big)\big\}_{j=1}^{M_{i+1}}$
   \STATE Sample a corresponding mini batch of utterance $y_i$.
   \STATE $y_i \sim P_{\theta_G}\big(y_i|,z_i,\boldsymbol{x_i}\big)$ 
   \ENDFOR
   \STATE Compute the discriminator accuracy $D_{acc}$ over $N-1$ utterances $\{y_i\}_{i=1}^{N-1}$ and $\{x_{i+1}\}_{i=1}^{N-1}$
   \IF{$D_{acc} < acc_{D_{th}}$}
   \STATE Update $\theta_D$ with gradient of the discriminator loss.
   \STATE $\sum\limits_{i}[\nabla_{\theta_D}\log{}D(\boldsymbol{h_i}, x_{i+1}) + \nabla_{\theta_D}log\big(1-D(\boldsymbol{h_i}, y_i)\big)]$
   \ENDIF
   \IF{$D_{acc} < acc_{G_{th}}$}
   \STATE Update $\theta_G$ with the generator's MLE loss only.
   \STATE $\sum\limits_{i}[\nabla_{\theta_G}\log{}P_{\theta_G}\big(y_i|,z_i,\boldsymbol{x_i}\big)]$
   \ELSE
   \STATE Update $\theta_G$ with both adversarial and MLE losses.
   \STATE $\sum\limits_{i}[\lambda_{G}\nabla_{\theta_G}\log{}D(\boldsymbol{h_i}, y_i) + 
                  \lambda_{M}\nabla_{\theta_G}\log{}P_{\theta_G}\big(y_i|,z_i,\boldsymbol{x_i}\big)]$
   \ENDIF
   \ENDFOR

\caption{Adversarial Learning of hredGAN}
\end{algorithmic}
\end{algorithm}

\section{Training of hredGAN}
We trained both the generator and the discriminator simultaneously as highlighted in Algorithm \ref{alg:hredgan} 
with $\lambda_G = \lambda_M = 1$. GAN training is prone to instability due to competition between the generator and the discriminator.  
Therefore, parameter updates are conditioned on the discriminator performance \citep{Lamb2016}.

\textbf{The generator} consists of four RNNs with different parameters, that is, $aRNN, eRNN, cRNN$, and $dRNN$. 
$aRNN$ and $eRNN$ are both bidirectional, while $cRNN$ and $dRNN$ are unidirectional. Each RNN has 3 layers, 
and the hidden state size is 512. The $dRNN$ and $aRNN$ are connected using an additive attention mechanism \citep{Bahdanau2015}.

\textbf{The discriminator} shares $aRNN, eRNN$, and $cRNN$ with the generator. $D_{RNN}$ is a stacked 
bidirectional RNN with 3 layers and a hidden state size of 512. 
The $cRNN$ states are used to initialize the states of $D_{RNN}$. 
The output of both the forward and the backward cells for each word are 
concatenated and passed to a fully-connected layer with binary output. The output is 
the probability that the word is from the ground truth given the past and future words of the sequence.

\textbf{Others:} All RNNs used are gated recurrent unit (GRU) cells \citep{Cho2014}.
The word embedding size is 512 and shared between the generator and the discriminator.
The initial learning rate is $0.5$ with decay rate factor of $0.99$, applied when 
the adversarial loss has increased over two iterations. We use a batch size of 64 and clip gradients 
around $5.0$. As in \citet{Lamb2016}, we find $acc_{D_{th}} = 0.99$ and $acc_{G_{th}} = 0.75$ to suffice.
All parameters are initialized with Xavier uniform random initialization \citep{Glorot2010}. 
The vocabulary size $V$ is $50,000$. Due to the large vocabulary size,
we use sampled softmax loss \citep{Jean2015} for MLE loss to expedite the training process. However, we use full softmax
for evaluation. The model is trained end-to-end using the stochastic gradient descent algorithm.  
\section{Experiments and Results}
We consider the task of generating dialogue responses conditioned on the dialogue history and the current input utterance.
We compare the proposed hredGAN model against some alternatives on publicly available datasets. 

\subsection{Datasets}
\textbf{Movie Triples Corpus} (MTC) dataset \citep{Serban2016}. This dataset was derived from the 
\textit{Movie-DiC} dataset by \citet{Banchs2012}. Although this dataset spans a wide range of topics with 
few spelling mistakes, its small size of only about 240,000 dialogue triples makes it difficult to train a dialogue 
model, as pointed out by \citet{Serban2016}. We thought that this scenario would really benefit from the 
proposed adversarial generation.

\textbf{Ubuntu Dialogue Corpus} (UDC) dataset \citep{Serban2017}. This dataset was extracted from the Ubuntu 
Relay Chat Channel. Although the topics in the dataset are not as diverse as in the MTC, the dataset is very large, containing 
about 1.85 million conversations with an average of 5 utterances per conversation.

We split both MTC and UDC into training, validation, and test sets, using 90\%, 5\%, and 5\% proportions, respectively. 
We performed minimal preprocessing of the datasets by replacing all words except the top 50,000 most frequent words by an 
\textit{UNK} symbol.     



\begin{table*}[ht]
\vspace{5pt}
\begin{center}
\begin{small}
\setlength\tabcolsep{4.0pt}
\begin{tabular}{l|ll|cccc|c}
\toprule
\multirow{2}{*}{Model}    & \multicolumn{2}{c|}{Teacher Forcing} & \multicolumn{4}{c|}{Autoregression} & Human \\    
 & Perplexity & $-logD(G(.))$ & BLEU-2 & ROUGE-2 & DISTINCT-1/2 & NASL & Evaluation\\
\midrule
\textbf{MTC} & & & & & & & \\
HRED      & 31.92/36.00  & NA           & 0.0474   & 0.0384 & 0.0026/0.0056 & 0.535 & 0.2560 $\pm$ 0.0977 \\
VHRED     & 42.61/44.97  & NA           & 0.0606   & 0.1181 & 0.0048/0.0163 & 0.831 & 0.3909 $\pm$ 0.0240 \\
hredGAN\_u & \textbf{23.57/23.54} & \textbf{6.85/6.81}    & 0.0493   & 0.2416 & 0.0167/0.1306 & 0.884 & 0.5582 $\pm$ 0.0118\\
hredGAN\_w & 24.20/24.14 & 13.35/13.40  & \textbf{0.0613}   & \textbf{0.3244} & \textbf{0.0179/0.1720} & \textbf{1.540} & \textbf{0.7869 $\pm$ 0.1148}\\
\midrule
\textbf{UDC} & & & & & & & \\
HRED       & 69.39/86.40  & NA          & 0.0177  & 0.0483  & 0.0203/0.0466 & 0.892 & 0.3475  $\pm$ 0.1062 \\
VHRED      & 98.50/105.20 & NA          & 0.0171  & 0.0855  & 0.0297/0.0890 & 0.873 & 0.4046 $\pm$ 0.0188\\
hredGAN\_u & 56.82/57.32  & 10.09/10.08 & 0.0137  & 0.0716  & 0.0260/0.0847 & \textbf{1.379} & 0.6133 $\pm$ 0.0361 \\
hredGAN\_w & \textbf{47.73/48.18}  & \textbf{8.37/8.36}   & \textbf{0.0216}  & \textbf{0.1168}  & \textbf{0.0516/0.1821} & 1.098 & \textbf{0.6905 $\pm$ 0.0706}\\
\bottomrule
\end{tabular}
\end{small}
\end{center}
\caption{Generator Performance Evaluation}
\label{tb:perp}
\end{table*}

\subsection{Evaluation Metrics}
Accurate evaluation of dialogue models is still an open challenge. In this paper, we employ both automatic and human evaluations. 
\subsubsection{Automatic Evaluation}
We employed some of the automatic evaluation metrics that are used in probabilistic language and dialogue models, 
and statistical machine translation. Although these metrics may not correlate well 
with human judgment of dialogue responses \citep{Liu2016}, they provide a good baseline for comparing dialogue model performance.

\textbf{Perplexity} - For a model with parameter $\theta$, we define perplexity as:
\begin{equation}\abovedisplayskip=0pt \belowdisplayskip=0pt
exp\bigg[-\frac{1}{N_W}\sum_{k=1}^K log~P_{\theta}(y_1, y_2,\dots, y_{N_k-1})\bigg]
\end{equation}
where $K$ is the number of conversations in the dataset, $N_k$ is the number of utterances in conversation $k$, 
and $N_W$ is the total number of word tokens in the entire dataset. The lower the perplexity, the better. The 
perplexity measures the likelihood of generating the ground truth given the model parameters. While a generative model 
can generate a diversity of responses, it should still assign a high probability to the ground truth utterance. 

\textbf{BLEU} - The BLEU score \citep{Papineni2002} provides a measure of overlap between the generated response (candidate) 
and the ground truth (reference) using a modified n-gram precision. According to Liu et.~al. \citep{Liu2016}, BLEU-2 score 
is fairly correlated with human judgment for non-technical dialogue (such as MTC).

\textbf{ROUGE} - The ROUGE score \citep{Lin2014} is similar to BLEU but it is recall-oriented instead. 
It is used for automatic evaluation of text summarization and machine translation. To compliment the BLEU score, we use ROUGE-N 
with $N=2$ for our evaluation.

\textbf{Distinct n-gram} - This is the fraction of unique n-grams in the generated responses and it provides a measure of diversity. 
Models with higher a number of distinct n-grams tend to produce more diverse responses \citep{Li2016}. For our evaluation, we use 1- and 2- grams.

\textbf{Normalized Average Sequence Length (NASL)} - This measures the average number of words in model-generated responses normalized 
by the average number of words in the ground truth.

\subsubsection{Human Evaluation}
For human evaluation, we follow a similar setup as \citet{Li2016}, employing crowd-sourced judges to evaluate a random selection of 200 samples.
We presented both the multi-turn context and the generated responses from the models to 3 judges and asked them to rank the general response quality in terms 
of relevance and informativeness. For $N$ models, the model with the lowest quality is assigned a score 0 and the highest is assigned a score N-1. Ties are 
not allowed. The scores are normalized between 0 and 1 and averaged over the total number of samples and judges. For each model, we also estimated the per sample score variance between judges and then averaged over the number of samples, i.e., sum of variances divided by the square of number of samples (assuming sample independence). The square root of result is reported as the standard error of the human judgment for the model.

\subsection{Baseline}

We compare the performance of our model to (V)HRED \citep{Serban2016,Serban2017}, since they are the closest to our approach in 
implementation and are the current state of the art in open-domain dialogue models.
HRED is very similar to our proposed generator, but without the input utterance attention and noise samples.
VHRED introduces a latent variable to the HRED between the \textit{cRNN} and the \textit{dRNN} and was trained using the variational lower 
bound on the log-likelihood. The VHRED can generate multiple responses per context like hredGAN, but it has no specific 
criteria for selecting the best response.

The HRED and VHRED models are both trained using the Theano-based implementation obtained from \url{https://github.com/julianser/hed-dlg-truncated}. 
The training and validation sets used for UDC and MTC dataset were obtained directly from the authors\footnote{UDC was obtained from
\url{http://www.iulianserban.com/Files/UbuntuDialogueCorpus.zip}, and the link to MTC was obtained privately.} of (V)HRED.
For model comparison, we use a test set that is disjoint from the training and validation sets.

\subsection{Results}
We have two variants of hredGAN based on the noise injection approach, i.e., hredGAN with utterance-level 
(\textit{hredGAN\_u}) and word-level (\textit{hredGAN\_w}) noise injections.

We compare the performance of these two variants with HRED and VHRED models. 

\textbf{Perplexity}: The average perplexity per word performance of all the 
four models on MTC and UDC datasets (validation/test) are reported in the first column on Table \ref{tb:perp}.
The table indicates that both variants of the hredGAN model perform better 
than the HRED and VHRED models in terms of the perplexity measure. 
However, using the adversarial loss criterion (Eq. \eqref{eq:mlegan}), the hredGAN\_u model performs 
better on MTC and worse on UDC. Note that, for this experiment, we run all models in teacher forcing mode.

\textbf{Generation Hyperparameter}: For adversarial generation, 
we perform a linear search for $\alpha$ between 1 and 20 at an increment 
of 1 using Eq. \eqref{eq:adv_gen}, with sample size $L=64$, on validation sets with models run in autoregression. 
The optimum values of $\alpha$ for hredGAN\_u and hredGAN\_w  for UDC are $7.0$ and $9.0$ respectively. The values for MTC 
are not convex, probably due to small size of the dataset, so we use the same $\alpha$ values as UDC. We however note that 
for both datasets, any integer value between 3 and 10 (inclusive) works well in practice.

\iftrue
\begin{table}[t] 
\begin{center}
\begin{tiny}
\setlength\tabcolsep{2.0pt}
\begin{tabularx}{\linewidth}{ llX }
\toprule
Item & $D(G(.))$ & Utterance \\
\midrule
\textbf{MTC}\\
Context\_0& NA & perhaps $<$person$>$ had a word with the man upstairs .\\
Context\_1& NA & a word ? i ' m sure by now he ' s engineered a hostile takeover . \\
Response\_0& 0.996& $<$person$>$ , i know what you ' re saying , $<$person$>$ , that ' s not what i ' m saying . \\
Response\_1& 0.991& $<$person$>$ , i know . i was just about to help the guy . \\
Response\_2& 0.315& $<$person$>$ , i ' m sorry . \\
Response\_3& 0.203& $<$person$>$ , i ' m a little out . \\
\midrule
Context\_0& NA & says he wanted food . $<$person$>$ . he wanted the gold .\\
Context\_1& NA & how ' s he going to want the gold ? he couldn ' t even know we had it . \\
Response\_0& 0.998& $<$person$>$ , i know . but it ' s not him , it ' s the only way he ' s got it all figured \\
Response\_1& 0.981& $<$person$>$ , i know . but i have to tell you . these things are really stupid and you think i was wrong ? \\
Response\_2& 0.690& $<$person$>$ , i ' m sure he did . \\
Response\_3& 0.314& $<$person$>$ , i ' m not sure . \\
\toprule
\textbf{UDC}\\
Context\_0& NA & The netboot one is suppose to download packages from the net. \\
Context\_1& NA & like the ones to be installed? or the installed to be run? \\
Response\_0& 0.993& you don ' t need to install the whole system , just install the ubuntu installer \\
Response\_1& 0.952& you can install the ubuntu installer from the ubuntu menu \\
Response\_2& 0.749& I ' m not sure , I don ' t know . \\
Response\_3& 0.184& you can \\
Response\_4& 0.003& you a . 1 . . . . . . . . . . . . . . . . . . . . . \\
\midrule
Context\_0& NA & DJones: update manager won't detect 12.04.1 as a new version if you are already running 12.04, because 12.04.1 = 12.04 + lots of package updates \\
Response\_0& 0.991& did you try a clean install of the latest version ? \\
Response\_1& 0.981& try installing the latest \_UNK and see if it works \\
Response\_2& 0.615& I ' m not sure you have a problem . . . \\
Response\_3& 0.191& try sudo apt-get remove the package that is not installed \\
Response\_4& 0.002& try the \_UNK . 1 . 1 . 1 . 1 . 1 . \_UNK . deb \\
\bottomrule
\end{tabularx}
\end{tiny}
\end{center}
\caption{Example of Discriminator Ranking with hredGAN\_w}
\label{tb:conv}
\end{table}
\begin{table}[t]
\begin{center}
\begin{tiny}
\setlength\tabcolsep{1.0pt} 
\begin{tabularx}{\linewidth}{ l X }
\toprule
\textbf{Movie} \\
\toprule
\textbf{Context\_0}& perhaps $<$person$>$ had a word with the man upstairs .\\
\textbf{HRED}& i ' m sorry , $<$person$>$ .\\
\textbf{VHRED}& $<$person$<$ ' t do it . \\
\textbf{hredGAN\_w}& $<$person$>$ , i ' m not gon na be okay . i just don ' t think it ' s a good idea .\\
\textbf{hredGAN\_u}&  $<$person$>$ , i cause it was all the way to you . \\
\midrule 
\textbf{Context\_1}& a word ? i ' m sure by now he ' s engineered a hostile takeover . \\
\textbf{HRED}&  i ' m sorry , $<$person$>$ . \\
\textbf{VHRED}& $<$person$>$ ' t do it . \\
\textbf{hredGAN\_w}& $<$person$>$ , i know what you ' re saying , $<$person$>$ , that ' s not what i ' m saying . \\
\textbf{hredGAN\_u}&  $<$person$>$ , i have to tell you i ' m not the same . \\
\toprule
\textbf{Context\_0}& $<$person$>$ ' s .\\
\textbf{HRED}& i ' m sorry , $<$person$>$ .\\
\textbf{VHRED}& $<$person$>$ ' t do it . \\
\textbf{hredGAN\_w}& $<$person$>$ , i ' m sure it ' s the only thing i ever imagined . \\
\textbf{hredGAN\_u}&  $<$person$>$ , i ' ll be right there . \\
\midrule
\textbf{Context\_1}& can i ask you something about her ? \\
\textbf{HRED}&  i don ' t know . \\
\textbf{VHRED}:& you ' re not going to be here . \\
\textbf{hredGAN\_w}& i don ' t know . i think i ' ve heard of it . \\
\textbf{hredGAN\_u}&  i ' m sorry , i can ' t .\\
\toprule
\textbf{Ubuntu} \\
\toprule
\textbf{Context\_0}& this is Dacs i am in i used 'sudo statx and it let me in\\
\textbf{HRED}&  what 's the problem ? \\
\textbf{VHRED}&  I don't know what you want to do . if you want to do it , you need to do it to do it in the terminal ? \\
\textbf{hredGAN\_w}& you don ' t have to do anything , just type in the terminal and it should be to find the right device \\
\textbf{hredGAN\_u}&  you can have a look at the output of the command . . .\\
\bottomrule
\textbf{Context\_1}& good deal... cat /etc/X11/default-display-manager\\
\textbf{HRED}&   what 's the problem ? \\
\textbf{VHRED}&  do you know what you want to do ? \\
\textbf{hredGAN\_w }& I ' m trying to figure a command that I can find to find out the file that I can find in the file" \\
\textbf{hredGAN\_u}& I don ' t see the point , but I ' m not sure how to do that . \\
\midrule
\textbf{Context\_2}& /usr/sbin/lightdm http://paste.ubuntu.com/1286224/ $<$---- my /etc/X11/xorg.conf\\
\textbf{HRED}&   what 's the problem ? \\
\textbf{VHRED}&  is there a way to do that in the terminal ? \\
\textbf{hredGAN\_w }& did you just type \textbackslash" sudo mount -a \textbackslash" ? \\
\textbf{hredGAN\_u}& i have no idea , i just installed ubuntu and i have no idea how to do that \\
\bottomrule
\end{tabularx}
\end{tiny}
\end{center}
\caption{ Sample responses of HRED, VHRED and hredGAN.}
\label{tb:samples}
\end{table}
\fi

\textbf{Quantitative Generator Performance}: We run autoregressive inference for all the models 
(using optimum $\alpha$ values for hredGAN models and selecting the best of $L=64$ responses using a discriminator)
with dialogue contexts from a unique test set.
Also, we compute the average BLEU-2, ROUGE-2(f1), Distinct(1/2), and normalized average sequence length (NASL) scores for each model 
and summarize the results in the middle of Table \ref{tb:perp}. Distinct(1/2) largely agrees with the perplexity score.
Most scores, similar to the perplexity, indicate that hredGAN models perform better than (V)HRED on both datasets. 
However, on the UDC ROUGE and MTC BLEU, VHRED scores slightly better than hredGAN\_u but still worse than hredGAN\_w.

A good dialogue model should find the right balance between precision (BLEU) and diversity. 
We strongly believe that our adversarial approach is better suited to solving this problem. 
As hredGAN generators explore diversity, the discriminator ranking gives 
hredGAN an edge over (V)HRED because it helps detect responses that are out of 
context and the natural language structure (Table \ref{tb:conv}). 
Also, the ROGUE(f1) performance indicates that hredGAN\_w strikes a better balance between precision (BLEU) and diversity 
than the rest of the models. This is also obvious from the quality of generated responses.

\textbf{Qualitative Generator Performance:} The results of the human evaluation are reported in the last 
column of Table \ref{tb:perp}. The human evaluation agrees largely with the automatic evaluation. hredGAN\_w 
performs best on both datasets although the gap is more on the MTC than on the UTC. 
This implies that the improvement of HRED with adversarial generation is better than with variational generation (VHRED).
In addition, looking at the actual samples from the generator outputs in Table \ref{tb:samples} 
shows that hredGAN, especially hredGAN\_w, performs better than (V)HRED. 
While other models produce short and generic utterances, hredGAN\_w mostly yields informative responses.
For example, in the first dialogue in Table \ref{tb:samples}, when the speaker is sarcastic about ``the man upstairs'',
hredGAN\_w responds with the most coherent utterance with respect to the dialogue history. We see similar behavior across 
other samples. We also note that although hredGAN\_u's responses are the longest on Ubuntu (in line with the NASL score),   
the responses are less informative compared to hredGAN\_w resulting in a lower human evaluation score. 
We reckon this might be due to a mismatch between utterance-level 
noise and word-level discrimination or lack of capacity to capture the data distribution using single noise distribution. 
We hope to investigate this further in the future.

\textbf{Discriminator Performance:}
Although only hredGAN uses a discriminator, the observed discriminator behavior is interesting.
We observe that the discriminator score is generally reasonable with longer, 
more informative and more persona-related responses receiving higher scores as shown in Table \ref{tb:conv}. 
It worth to note that this behavior, although similar to the behavior of a human judge is learned without supervision.
Moreover, the discriminator seems to have learned to assign an average score to more frequent or generic responses such as 
``I don't know,'' ``I'm not sure,'' and so on, and high score to rarer answers. That's why we sample a modified noise 
distribution during inference so that the generator can produce rarer utterances that will be scored high by the 
discriminator.

\section{Conclusion and Future Work}

In this paper, we have introduced an adversarial learning approach that addresses response diversity and 
control of generator outputs, using an HRED-derived generator and discriminator. 
The proposed system outperforms existing state-of-the-art (V)HRED 
models for generating responses in multi-turn dialogue with respect to automatic and human evaluations.
The performance improvement of the adversarial generation (hredGAN) over the variational generation (VHRED) comes from the combination of adversarial 
training and inference which helps to address the lack of diversity and contextual relevance in maximum likelihood based generative 
dialogue models. 
Our analysis also concludes that the word-level noise injection seems to perform better in general.


\bibliography{acl2019_hred_gan}
\bibliographystyle{acl_natbib}

\appendix

\begin{table*}[t]
\begin{center}
\begin{small}
\setlength\tabcolsep{4.0pt}
\begin{tabular}{l|c|cccc}
\toprule
\multirow{2}{*}{Model}    & \multicolumn{1}{c|}{Teacher Forcing} & \multicolumn{4}{c}{Autoregression}\\    
 & Perplexity & BLEU-2 & ROUGE-2 & DISTINCT-1/2 & NASL\\
\midrule
\textbf{MTC} & & & & & \\
HRED      & 31.92/36.00            & 0.0474   & 0.0384 & 0.0026/0.0056 & 0.535 \\
HRED+Attn & 18.70/19.02            & 0.0425   & 0.2239 & 0.0397/0.1567 & 0.527\\
hredGAN\_no\_noise   & 18.93/19.19            & 0.0355   & 0.1839 & 0.0272/0.0978 & 0.471\\

\midrule
\textbf{UDC} & & & & & \\
HRED       & 69.39/86.40           & 0.0177  & 0.0483  & 0.0203/0.0466 & 0.892\\
HRED+Attn  & 43.43/43.92           & 0.0140  & 0.0720  & 0.0473/0.1262 & 0.760 \\
hredGAN\_no\_noise    & 43.48/44.04           & 0.0123  & 0.0827  & 0.0398/0.1147 & 0.908 \\
\bottomrule
\end{tabular}
\end{small}
\end{center}
\caption{Generator Performance: HRED, HRED+Attn and hredGAN without noise}
\label{tb:perp_attn}
\end{table*}

\section{Ablation Experiments}
Before proposing the above adversarial learning framework for multi-turn dialogue, we carried out some experiments.

\subsection{Generator:}
We consider two main factors here, i.e., addition of an attention memory and injection of Gaussian noise into the generator input.
\subsubsection{Addition of Attention Memory}
First, we noted that by adding an additional attention memory to the HRED generator, we improved
the test set perplexity score by more than 12 and 25 points on the MTC and UDC respectively as shown in Table \ref{tb:perp_attn}. 
The addition of attention also shows strong performance at autoregressive inference across multiple metrics as well as an observed improvement in response quality. Hence the decision for the modified HRED generator.   

\subsubsection{Injection of Noise}
Before injecting noise into the generator, we first train hredGAN without noise. The result is also reported in \ref{tb:perp_attn}. 
We observe accelerated generator training but without an appreciable improvement in performance. It seems the discrimination task is very easy 
since there is no stochasticity in the generator output. Therefore, the adversarial feedback does not meaningfully impact the generator weight update.

Finally, we also notice that even with noise injection, there is no appreciable improvement in the auto-regressive performance if we sample with $L=1$ 
even though the perplexity is higher. However, as we increase $L$, producing $L$ responses per turn, the discriminator's adversarial selection gives a better 
performance as reported in Table 1.

Therefore, we conclude that the combination of adversarial training and adversarial inference helps to address the lack of diversity and contextual relevance observed in the generated responses.

\subsection{Discriminator:}
Before deciding on the word-level discrimination, we experimented with utterance-level discrimination. 
The utterance-level discriminator trains very quickly but it leads to mostly 
generic responses from the generator. We also note that utterance-level discriminator scores are mostly extreme (i.e., either 
low or high). Since we had used a convolutional neural network discriminator \citep{Yu2017} in our experiments, 
we hope to investigate this further with other architectures. 


\subsection{Adversarial Training:}
Lastly, we also tried a basic policy gradient approach \citep{Li2017}, where word-level discriminator score is used 
as a reward for each generated word token, but this leads to training instability. This is probably due to the instability of 
Monte Carlo sampling over a large vocabulary size. We believe this might improve with other sampling methods such as 
importance sampling and hope to investigate this further in the future.

\iftrue
\clearpage
\begin{table*}[t]

\begin{minipage}[t]{0.48\textwidth}
\vspace{-8pt}
\begin{center}
\begin{tiny}
\begin{tabularx}{\linewidth}{ l X }
\toprule
\textbf{Model} & \textbf{Response} \\
\toprule
\textbf{Context\_0}& $<$person$>$ i ' ve raised my rate . i might even cut my sessions to forty minutes .\\
\textbf{HRED}& i ' m sorry , $<$person$>$ .\\
\textbf{VHRED}& $<$person$>$, $<$person$>$, $<$person$>$, $<$person$>$ . \\
\textbf{hredGAN\_w}& $<$person$>$ , i ' m sorry , but i ' m not sure if you ' re really a very good guy . \\
\textbf{hredGAN\_u}&  i ' ll never see them again , $<$person$>$ . \\
\midrule
\textbf{Context\_0}& i ' m going to handle this exactly the way your father would have handled this . i ' ve got it solved .\\
\textbf{Context\_1}& you ' ve got it solved ? where is she ?\\
\textbf{HRED}& i don ' t know .\\
\textbf{VHRED}& $<$person$>$, , i ' m sorry. \\
\textbf{hredGAN\_w}& to the end of a beautiful woman who will be very happy . \\
\textbf{hredGAN\_u}& i ' m sorry , i just didn ' t know -- \\
\midrule 
\textbf{Context\_0}& of course not ! <person> ' s like a daughter to me . her father and i taught at university together . a good man , but he never listened .\\
\textbf{HRED}& i ' m sorry , $<$person$>$. \\
\textbf{VHRED}& $<$person$>$, mr . $<$person$>$ . \\
\textbf{hredGAN\_w}& $<$person$>$ , i know it ' s been all six years since i told him about you . \\
\textbf{hredGAN\_u}& i ' m sure you didn ' t , but i would have done a little more . \\
\midrule 
\textbf{Context\_0}& of course not ! <person> ' s like a daughter to me . her father and i taught at university together . a good man , but he never listened .\\
\textbf{Context\_1}& neither did i .\\
\textbf{HRED}& i ' m sorry , $<$person$>$. \\
\textbf{VHRED}& $<$person$>$ , you ' re a $<$unk$>$ . \\
\textbf{hredGAN\_w}& $<$person$>$, i know you ' re think about it . think it ' ll be time , right ?\\
\textbf{hredGAN\_u}& i ' ll be the first one . if it did you , you ' ll be it .\\
\midrule
\textbf{Context\_0}& come on , $<$person$>$ , we could have a great week up in vermont .\\
\textbf{Context\_1}& i see what you mean . but wednesday we ' ve got to be back in new york .\\
\textbf{HRED}& $<$person$>$ .\\
\textbf{VHRED}& $<$person$>$ , i don ' t know what you ' re going to do . \\
\textbf{hredGAN\_w}& $<$person$>$ , i know . i ' m kind of busy here .\\
\textbf{hredGAN\_u}& i ' m sorry , i can ' t . \\
\midrule
\textbf{Context\_0}& isabel ? \\
\textbf{Context\_1}& hi . \\
\textbf{HRED}& that ' s what i ' m talking about . \\
\textbf{VHRED}:& you ' re not going to be a $<$unk$>$ , $<$person$>$ . \\
\textbf{hredGAN\_w}& $<$person$>$ ? \\
\textbf{hredGAN\_u}& what do you want to know ?\\
\midrule
\textbf{Context\_0}& you understand mr . $<$person$>$ is constantly barraged with nuisance litigation and irs audits . \\
\textbf{Context\_1}& of course . \\
\textbf{HRED}& that ' s right , $<$person$>$ . \\
\textbf{VHRED}:& $<$person$>$ , you ' re going to be a $<$unk$>$ of the $<$unk$>$ . \\
\textbf{hredGAN\_w}& $<$person$>$ , i ' m sure , but you ' re probably still in to the new $<$person$>$ ' s house , huh ? \\
\textbf{hredGAN\_u}& you ' re gon na be a little boy , and i ' m not gon na get out !\\
\midrule
\textbf{Context\_0}& $<$person$>$ ' t go into my room , she ' s asleep . \\
\textbf{Context\_1}& $<$person$>$ , but the nanny , mrs . $<$person$>$ , she wants to see her . \\
\textbf{HRED}& i ' m sorry , $<$person$>$ . \\
\textbf{VHRED}:& $<$person$>$ , mr . $<$person$>$ .\\
\textbf{hredGAN\_w}& $<$person$>$ , i know . but she ' s never seen me anyway .\\
\textbf{hredGAN\_u}& i ' m sorry i didn ' t know what to say .\\
\midrule
\textbf{Context\_0}& what was that for ?\\
\textbf{Context\_1}& i ' d forgotten how exciting it is working together .\\
\textbf{HRED}& i ' m sorry , $<$person$>$ . \\
\textbf{VHRED}:& $<$person$>$ , $<$person$>$ . i don ' t know .\\
\textbf{hredGAN\_w}& i ' m sorry , $<$person$>$ . so many people are so interested , but you ' re a very nice guy .\\
\textbf{hredGAN\_u}& i ' ve always been so sure to $<$person$>$ .\\
\midrule
\textbf{Context\_0}& a teacher ' s .\\
\textbf{Context\_1}& oh my god .\\
\textbf{HRED}& i ' m sorry , $<$person$>$ . \\
\textbf{VHRED}:& $<$person$>$ , $<$person$>$ ' t worry about it . i don ' t know what you ' re doing .\\
\textbf{hredGAN\_w}& i ' m sorry . i didn ' t mean to . i didn ' t mean to hurt you . i just wanted you\\
\textbf{hredGAN\_u}& $<$person$>$ , i don ' t want to be your friend .\\
\midrule
\textbf{Context\_0}& i took the initiative .\\
\textbf{Context\_1}& you haven ' t been given an assignment in this case .\\
\textbf{HRED}& $<$person$>$ . \\
\textbf{VHRED}:& $<$person$>$ , $<$person$>$ , $<$person$>$ , $<$person$>$ . i ' m not going to get out of the $<$unk$>$ .\\
\textbf{hredGAN\_w}& i ' m not sure . $< $continued\_utterance$ >$ i ' m not gon na do you this shit anyway , $<$person$>$ .\\
\textbf{hredGAN\_u}& $<$person$>$ , i guess i didn ' t .\\
\bottomrule
\end{tabularx}
\end{tiny}
\end{center}
\caption{Additional sample responses of HRED, VHRED and hredGAN on Movie.}
\label{tb:samples}
\end{minipage}%
\hspace{5pt}
\begin{minipage}[t]{0.5\textwidth}
\vspace{-8pt}
\begin{center}
\begin{tiny}
\begin{tabularx}{\linewidth}{ l X }
\toprule
\textbf{Model} & \textbf{Response} \\
\toprule
\textbf{Context\_0}& what's the command you are running to chroot and the error message you get in return?\\
\textbf{Context\_1}& i mountded the device sudo mount /dev/sda1 /mnt - that works - then sudo chroot /mnt -- error is no such file or command\\
\textbf{HRED}&  what 's the problem ? \\
\textbf{VHRED}&  is there a way to do that ?  what do you mean ? \\
\textbf{hredGAN\_w}& I have no idea , I just did a sudo fdisk and it said it was mounted \\
\textbf{hredGAN\_u}& I ' m trying to get the latest version of ubuntu , but I ' m getting a lot of problems .\\
\midrule
\textbf{Context\_0}& Hmm, have you checked that ICMP is being allowed all the way through and that the server responds to other ICMP requests?\\
\textbf{Context\_1}& now i can ping it (i had to enable the lzo compression on client) i cannot reach the remote subnet adding a rule in the firewall forward the lan to vpn\\
\textbf{HRED}&   what 's the problem ? \\
\textbf{VHRED}&  I don't know what you want to do \\
\textbf{hredGAN\_w }& I have no idea , I ' m just trying to get the ip address from the router to the router" \\
\textbf{hredGAN\_u}& I ' m not sure if that ' s the case , but I ' m not sure if it ' s a good idea to have a look at the output .\\
\midrule
\textbf{Context\_0}& The netboot one is suppose to download packages from the net. \\
\textbf{HRED}& I don't know what you want to do . \\
\textbf{VHRED}& how do I do that ? \\
\textbf{hredGAN\_w}& I ' m not sure of a command to do so I can just install the ubuntu . iso and burn a new iso \\
\textbf{hredGAN\_u}&  you can ' t have a look at the man page . . . \\
\midrule
\textbf{Context\_0}& The netboot one is suppose to download packages from the net. \\
\textbf{Context\_1}& like the ones to be installed? or the installed to be run? \\
\textbf{HRED}&  I don't know what the problem is . \\
\textbf{VHRED}& is there a way to install the package manager ? \\
\textbf{hredGAN\_w}& you don ' t need to install the whole system , just install the ubuntu installer \\
\textbf{hredGAN\_u}& I ' m trying to install a \_UNK version of ubuntu , but I ' m trying to install it on my laptop .\\
\midrule
\textbf{Context\_0}& You trying to mount it open with the password?\\
\textbf{HRED}&  I have to do that , but I have a problem with the same problem . \\
\textbf{VHRED}&  is there a way to do ? \\
\textbf{hredGAN\_w}& yes , I ' m not sure if it ' s mounted or not \\
\textbf{hredGAN\_u}& I ' m trying to install the latest version of Ubuntu , but I ' m running on a \_UNK .\\
\midrule
\textbf{Context\_0}& are you using lvm for your root partition?\\
\textbf{HRED}&  yes , but I have a problem with a new user , but I don't want to do that . \\
\textbf{VHRED}&  what is the best way to do that ? what is that ? \\
\textbf{hredGAN\_w}& yes , I ' m on a live system , I ' m trying to use the partition to boot from windows \\
\textbf{hredGAN\_u}& I ' m not sure if it ' s a good idea to have a backup of the system . . .\\
\midrule
\textbf{Context\_0}& are you using lvm for your root partition?\\
\textbf{Context\_1}& i have the same config working on a 10.04 box. It appears in 12.04 lvm is on upstart and multipath is not. i do use lvm for my root folder nifty, is there any way to work around it? have you heard of anyone posting a bug about this to have it looked at? darn.... well i hope they can fix this eventually i hope they keep initrd around and move everything to upstart. Upstart is not easy to setup for programs that don't come packaged for upstart.\\
\textbf{HRED}&   I don't know what the problem is , but I don't know what the problem is . ? \\
\textbf{VHRED}&  is there a way to do that ? or **unknown** ? \\
\textbf{hredGAN\_w }& you can also try the live cd and see if you can get it back up the live" \\
\textbf{hredGAN\_u}& I ' m trying to get my wireless working on my laptop , but I ' m getting to the point that I can be able to use the wireless card to get the wireless working .\\
\midrule
\textbf{Context\_0}& are you using lvm for your root partition?\\
\textbf{Context\_1}& i have the same config working on a 10.04 box. It appears in 12.04 lvm is on upstart and multipath is not. i do use lvm for my root folder nifty, is there any way to work around it? have you heard of anyone posting a bug about this to have it looked at? darn.... well i hope they can fix this eventually i hope they keep initrd around and move everything to upstart. Upstart is not easy to setup for programs that don't come packaged for upstart.\\
\textbf{Context\_2}& yes, because i cant supply it at boot time, i want the truecrypt drive to come up by itself without manual intervention\\
\textbf{HRED}&   what is the problem ? \\
\textbf{VHRED}&  what do you mean ? \\
\textbf{hredGAN\_w }& you can also mount a new one and put the mount command to the mount point" \\
\textbf{hredGAN\_u}& I ' m trying to get my sound working , but I ' m trying to get my sound working .\\
\bottomrule
\end{tabularx}
\end{tiny}
\end{center}
\caption{Additional sample responses of HRED, VHRED and hredGAN on Ubuntu.}
\label{tb:samples}
\end{minipage}
\end{table*}
\fi

\end{document}